\documentclass[english]{article}
\usepackage[T1]{fontenc}
\usepackage[latin9]{inputenc}
\usepackage{xcolor}
\usepackage{array}
\usepackage{booktabs}
\usepackage{units}
\usepackage{multirow}
\usepackage{amsmath}
\usepackage{graphicx}

\makeatletter

\providecommand{\tabularnewline}{\\}

\RequirePackage{color}
\RequirePackage{subfigure}
\RequirePackage{xspace}
\RequirePackage{eso-pic}

\makeatletter
\newcommand*{\etal}{%
    \@ifnextchar{.}%
        {\textit{et al}}%
        {\textit{et al.}\@\xspace}%
}
\makeatother

\date{}

\makeatother

\usepackage{babel}
\begin{document}
\title{Learning to Abstract and Predict Human Actions}
\author{Romero Morais$^{*}$, Vuong Le, Truyen Tran, Svetha Venkatesh\\
 Applied Artificial Intelligence Institute, Deakin University, Australia\\
 \texttt{\small{}\{ralmeidabaratad,vuong.le,truyen.tran,svetha.venkatesh\}@deakin.edu.au}}

\maketitle
\global\long\def\const{\text{const}}%

\global\long\def\Encoder{\text{Encoder}}%
\global\long\def\Decoder{\text{Anticipator}}%
\global\long\def\Transitioner{\text{Refresher}}%

\global\long\def\Model{\text{HERA}}%
\global\long\def\ModelName{\text{Hierarchical \ensuremath{\Encoder}-\ensuremath{\Transitioner}-\ensuremath{\Decoder}}}%

\begin{abstract}
Human activities are naturally structured as hierarchies unrolled
over time. For action prediction, temporal relations in event sequences
are widely exploited by current methods while their semantic coherence
across different levels of abstraction has not been well explored.
In this work we model the hierarchical structure of human activities
in videos and demonstrate the power of such structure in action prediction.
We propose Hierarchical Encoder-Refresher-Anticipator, a multi-level
neural machine that can learn the structure of human activities by
observing a partial hierarchy of events and roll-out such structure
into a future prediction in multiple levels of abstraction. We also
introduce a new coarse-to-fine action annotation on the \emph{Breakfast
Actions} videos to create a comprehensive, consistent, and cleanly
structured video hierarchical activity dataset. Through our experiments,
we examine and rethink the settings and metrics of activity prediction
tasks toward unbiased evaluation of prediction systems, and demonstrate
the role of hierarchical modeling toward reliable and detailed long-term
action forecasting.
\vspace{-2mm}
\end{abstract}

\section{Introduction}

An AI agent that shares the world with us needs to efficiently anticipate
human activities to be able to react to them. Moreover, the ability
to anticipate human activities is a strong indicator of the competency
in human behavior understanding by artificial intelligence systems.
While video action recognition \cite{carreira2017quo} and short-term
prediction \cite{ke2019time-conditioned} have made much progress,
reliable long-term anticipation of activities remains challenging
\cite{farha2019uncertainty-aware} as it requires deeper understanding
of the action patterns.

\begin{figure}
\centering{}\includegraphics[width=0.9\textwidth]{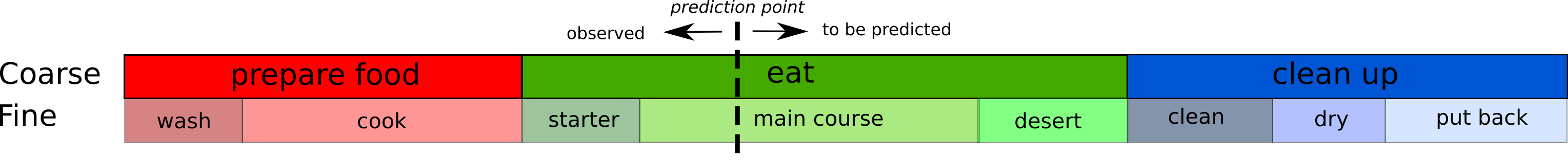}\caption{Illustration of a two-level structure of activity ``have dinner''
and a prediction task. \label{fig:activity-hierarchy-prediction}}
\vspace{-2mm}
\end{figure}

The most successful methods for activity prediction rely on modeling
the continuity of action sequences to estimate future occurrence by
neural networks \cite{farha2018when,ke2019time-conditioned}. However,
these networks only consider the sequential properties of the action
sequence which tends to fade and entice error accumulation in far-term.
This issue suggests exploring the abstract structure of actions that
spans over the whole undertaking of the task. One intuitive way to
approach this path is to follow the natural human planning process
that starts with high level tasks then proceeds to more refined sub-tasks
and detailed actions \cite{ajzen1985intentions}. An example of such
structure in an activity is shown in Fig.~\ref{fig:activity-hierarchy-prediction}.
Our quest is to build a neural machine that can learn to explore such
structures by observing a limited section of the video and extrapolate
the activity structure into the future for action prediction.

We realize this vision by designing a neural architecture called Hierarchical
Encoder-Refresher-Anticipator ($\Model$) for activity prediction.
$\Model$ consists of three sub-networks that consecutively encode
the past, refresh the transitional states, and decode the future until
the end of the overall task. The specialty of these networks is that
their layers represent semantic levels of the activity hierarchy,
from abstract to detail. Each of them operates on its own clock while
sending its state to parent layer and laying out plans for its children.
This model can be trained end-to-end and learn to explore and predict
the hierarchical structure of new video sequences. We demonstrate
the effectiveness of $\Model$ in improved long-term predictions,
increased reliability in predicting unfinished activities, and effective
predictions of activities at different levels of granularity. 

To promote further research in hierarchical activity structures, we
also introduce a new hierarchical action annotation to the popular
Breakfast Actions dataset \cite{kuehne2014language}. These annotations
contain two-level action labels that are carefully designed to reflect
the clean hierarchy of actions following natural human planning. In
numbers, it includes 25,537 annotations in two levels on 1,717 videos
spanning 77 hours. Once publicly released, this dataset will provide
a key data source to support advancing deep understanding into human
behaviors with potential applications in detection, segmentation and
prediction.

\vspace{-2mm}

\section{Related work}

For prediction of actions in videos, the most popular approach is
to predict the temporal action segments, by jointly predicting the
action labels and their lengths \cite{mahmud2017joint}. Recent advances
in this front include Farha \etal \cite{farha2018when} where random
prediction points are used with the RNN/CNN-like model. Moving away
from recurrent networks which tend to accumulate errors, Ke \etal
\cite{ke2019time-conditioned} used time point as the conditioning
factor in one-shot prediction approach with the trade-off in high
prediction cost and sparse predictions. While these methods work relatively
well in near-term, when the actions are predicted farther into the
future, uncertainty prevents them from having reliable results. Variational
methods manage uncertainty by using probabilistic modeling to achieve
more robust estimation of inter-arrival time \cite{mehrasa2019variational}
and action length \cite{farha2019uncertainty-aware}.

As an action is highly indicative of the next action, Miech \etal
\cite{miech2019leveraging} proposed a model that is a convex combination
of a ``predictive'' model and a ``transitional'' model. A memory-based
approach network was proposed by Gammulle \etal \cite{gammulle2019forecasting},
in which two streams with independent memories analyze visual and
label features to predict the next action. 

The hierarchy of activities can be considered in atomic scales where
small movements constitute an action \cite{lan2014hierarchical}.
Early works investigated the hierarchy of activity through layered
HMM \cite{duong2009efficient}, layered CRF \cite{Truyen:2006}, and
linguistic-like grammar \cite{qi2017predicting}. More recent works
favor neural networks due to their strong inductive properties \cite{farha2018when,ke2019time-conditioned}.
For hierarchy, Recurrent Neural Networks (RNN) can be stacked up,
but stacking ignores the multi-clock nature of a hierarchy unrolled
over time. In \cite{zhong2018time}, a hierarchical RNN with asynchronous
clocks was used to model the temporal point processes of activity
but the information only passes upward and multi-level semantics of
events are not explored. The idea of multi-clocks was also explored
by Hihi and Bengio \cite{hihi1996hierarchical} and Koutnik \etal
\cite{koutnik2014clockwork}. The drawback of these methods is that
the periods of the clock must be manually defined, which is not adaptive
to data structure at hand. Chung \etal \cite{chung2016hierarchical}
addressed this problem with a hierarchical multi-scale RNN (HM-RNN),
which automatically learns the latent hierarchical structure. This
idea has been extended with attention mechanism for action recognition
\cite{yan2018hierarchical}. Our hierarchical modeling shares the
structure exploration functionality with these works but is significantly
different in the ability to learn the semantic-rich structures where
layers of hierarchy are associated with levels of activity abstraction.
In particular, in comparison with Clock-work RNN (CW-RNN) \cite{koutnik2014clockwork},
$\Model$ shares the fact that units can update at different rates,
but $\Model$ is significantly different to CW-RNN in separating the
levels of RNN with distinctive associated semantics. $\Model$ also
allows RNN units to control their own clocks and their interactions
with other units.

\vspace{-2mm}

\section{Learning to abstract and predict human actions}

\subsection{Problem formulation\label{subsec:Problem-Formulation}}

We formalize an activity hierarchy $H$ of $L$ levels of a human
performing a task observable in a video as $H=\left\{ A^{l}\right\} _{l=1,2,\ldots,L}$
where each level $A^{l}$ is a sequence of indexed actions:
\begin{equation}
A^{l}=\left\{ \left(x_{k}^{l},d_{k}^{l}\right)\right\} _{k=1,2,\ldots,n_{l}}.
\end{equation}
Here, $x_{k}^{l}$ represents the label of $k$-th action at the $l$-th
level, $d_{k}^{l}$ is its relative duration calculated as its portion
of the parent activity, and $n_{l}$ indicates the number of actions
at level $l$. Each action $\left(x_{k}^{l},d_{k}^{l}\right)$ is
associated with a subsequence of finer actions at level $l+1$, and
the latter are called children actions of the former. Any children
subsequence is constrained to exclusively belong to only one parent
activity\footnote{Note that a child action label can be performed by multiple parents
at different parent times. See Sec.~\ref{subsec:Breakfast-Actions-Annotation}.}.

In the special case of a hierarchy with two levels, members of the
first level represent \emph{coarse activities}, and those at the second
level are called \emph{fine actions.} In this case, we will extend
the notation to use the level indices $c$ - for coarse and $f$ -
for fine in place of numeric indices $l=1$ and $l=2.$ An example
of a two-level hierarchy is shown in Fig.~\ref{fig:activity-hierarchy-prediction},
where for a \emph{task} of \texttt{\small{}<have-dinner>}, the first
coarse activity \texttt{\small{}<prepare-food>} contains three fine
actions as children.

Under this structure, the prediction problem is formed when the hierarchy
of activities is interrupted at a certain time $t^{*}$ indicating
the point where observation ends. At this time, at every level we
have finished events, unfinished events, and the task is to predict
events yet to start. The given observation includes the labels and
lengths of the finished events, and the labels and \emph{partial lengths}
of the unfinished ones. Thus the task boils down to estimating the
remaining lengths of the unfinished events, and all details of the
remaining events.

\subsection{$\protect\ModelName$}

\begin{figure*}
\centering{}\includegraphics[width=0.9\textwidth]{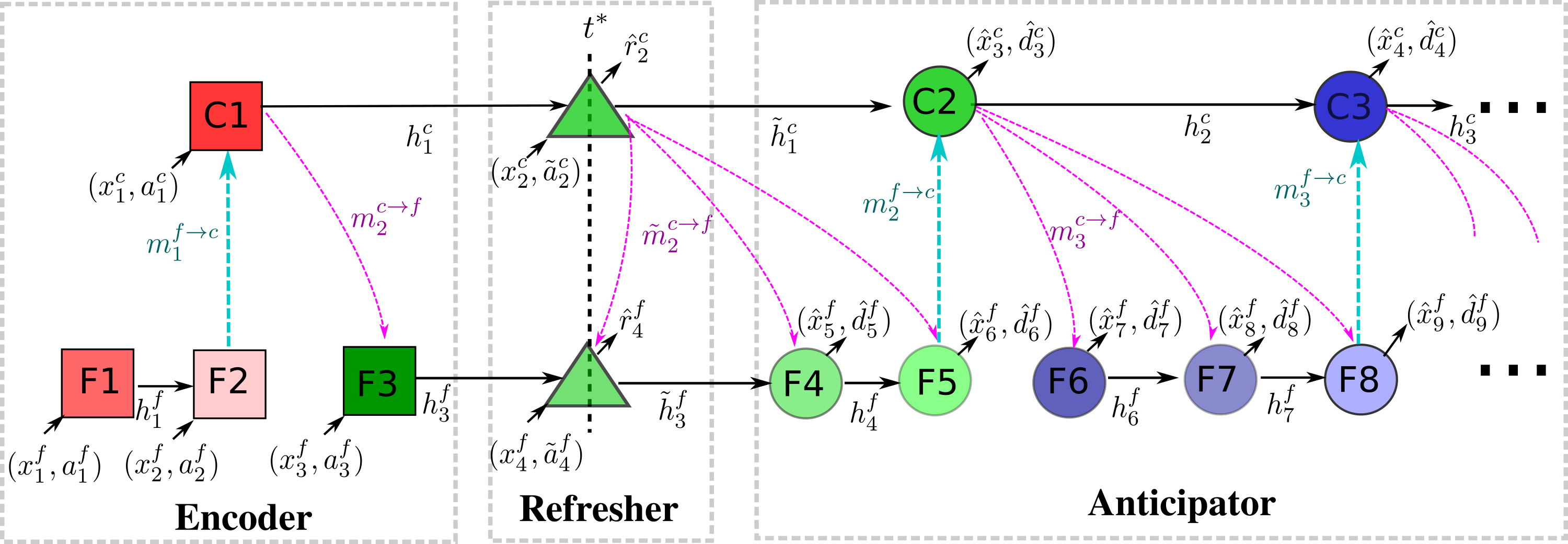}\caption{The $\protect\ModelName$ ($\protect\Model$) architecture realized
in a particular event sequence similar to the one in Fig.~\ref{fig:activity-hierarchy-prediction}.
Square blocks are $\protect\Encoder$ GRU cells, while triangles and
circles are those of Refresher and $\protect\Decoder$, respectively.
Color shades indicate cells processing different activity families,
e.g., the first coarse cell (\textcolor{red}{red} C1) and its two
children (\textcolor{red}{fading} \textcolor{brown}{red} F1 and F2)
process the first activity family $\{(x_{1}^{c},d_{1}^{c}),(x_{1}^{f},d_{1}^{f}),(x_{2}^{f},d_{2}^{f})\}$.
The prediction point $t^{*}$ happens at the middle of $(x_{2}^{c},d_{2}^{c})$
and $(x_{3}^{f},d_{3}^{f})$. Black arrows indicate recurrent links
while those in \textcolor{magenta}{pink} and \textcolor{cyan}{cyan}
are for downward and upward messages, respectively. For visual clarity,
optional prediction outputs of Encoder cell and feedback inputs of
Anticipator cell are omitted. \label{fig:method_architecture}}
\end{figure*}

We design $\Model$ (Fig.~\ref{fig:method_architecture}) to natively
handle the hierarchical structure of observation and extend such structure
to prediction. $\Model$ has three components: the $\Encoder$, the
$\Transitioner$, and the $\Decoder$. The $\Encoder$ creates a multi-level
representation of the observed events which is used by the $\Transitioner$
and $\Decoder$ to roll-out in a similar manner. The $\Encoder$ and
$\Decoder$ share the same hierarchical model design for cross-level
interaction which we detail next.

\paragraph{Modeling activity hierarchy. }

The $\Encoder$ and $\Decoder$ share an identical architecture of
two layers of recurrent neural units (RNN) which are chosen to be
based on Gated Recurrent Units (GRU) \cite{cho2014learning}. The
upper layer models the dynamics of coarse activities:\vspace{-2mm}
\begin{equation}
h_{i}^{c}=\textrm{GRU}\left(\left[(x_{i}^{c},a_{i}^{c}),m_{i}^{f\rightarrow c}\right],h_{i-1}^{c}\right).
\end{equation}
The first input to the unit includes a tuple of coarse label $x_{i}^{c}$
and accumulated duration $a_{i}^{c}=\sum_{k=1}^{i}d_{k}^{c}$. Both
$x_{i}^{c}$ and $a_{i}^{c}$ are encoded using a random embedding
matrix. At the $\Decoder$, these inputs are feedback from the previous
prediction step. The second input $m_{i}^{f\rightarrow c}$ is the
upward message that will be discussed later.

The lower layer is another RNN that is triggered to start following
the parent's operation:\vspace{-2mm}
\begin{equation}
h_{j}^{f}=\textrm{GRU}\left(\left[(x_{j}^{f},a_{j}^{f}),m_{i}^{c\rightarrow f}\right],h_{j-1}^{f}\right),
\end{equation}
where the proportional accumulated duration $a_{j}^{f}$ is calculated
within the parent activity.

By design, the two layers are asynchronous (i.e. the layers update
their hidden state independently and whenever fit) as coarse activities
happen sparser than fine actions. A key feature of $\Model$ is the
way it connects these two asynchronous concurrent processes in a consistent
hierarchy by using the cross-level messages. The downward message
$m_{i}^{c\rightarrow f}$ (pink arrows in Fig.\ref{fig:method_architecture})
provides instructions from the previous coarse cell to the current
fine cells. This message contains the previous coarse hidden state
$h_{i-1}^{c}$ and can optionally contain the parent's predicted label
$\hat{x}_{i}^{c}$. The upward message $m_{i}^{f\rightarrow c}$ (cyan
arrows) to a coarse node $i$ from its children contains the information
about the detail roll-out in the fine actions. It is implemented as
the hidden state of the last child.

\paragraph{Making predictions.}

At each step of both levels, the hidden state of the current cell
is used to infer the label and duration of the next action through
multi-layer perceptrons (MLP):\vspace{-2mm}
\begin{align}
\left(\hat{x}_{i+1}^{c},\hat{d}_{i+1}^{c}\right) & =\textrm{MLP}\left(h_{i}^{c}\right);\qquad\left(\hat{x}_{j+1}^{f},\hat{d}_{j+1}^{f}\right)=\textrm{MLP}\left(h_{j}^{f}\right)
\end{align}

For the $\Decoder$, these predictions are rolled out until accumulated
relative duration reaches 100\%. At the fine level, this means the
end of the parent coarse activity while at the coarse level it means
the overall task is finished. At the $\Encoder$, the predictions
are optionally used in training and are discarded in testing. 

\paragraph{Connecting the past and the present to the future.}

The connection between $\Encoder$ and $\Decoder$ happens at the
interruption point $t^{*}$, where the observed hierarchy ends and
prediction starts. If $t^{*}$ is well aligned with the action boundary,
we can simply pass the last hidden states and predictions of the $\Encoder$
to the $\Decoder$ at the corresponding levels. However, these coincidences
are rare; in most cases, the interruption happens at the middle of
an action and leaves trailing unfinished activity and actions at different
stages. 

To connect this gap, we design the $\Transitioner$ which consists
of a pair of connected MLP cells (Triangle blocks in Fig.~\ref{fig:method_architecture}).
The coarse Refresher cell gathers all available data and predicts
the remaining length $\hat{r}_{i^{*}}^{c}$ of interrupted coarse
activity $i^{*}$: \vspace{-2mm}
\begin{align}
\tilde{h}_{i^{*}-1}^{c} & =\textrm{MLP}\left(\left[h_{i^{*}-1}^{c},x_{i^{*}}^{c},\tilde{a}_{i^{*}}^{c},\tilde{d}_{i^{*}}^{c}\right]\right);\qquad\hat{r}_{i^{*}}^{c}=\textrm{MLP}\left(\tilde{h}_{i^{*}-1}^{c}\right),
\end{align}
where $\tilde{d}_{i^{*}}^{c}$ and $\tilde{a}_{i^{*}}^{c}$are unfinished
duration and accumulated duration, respectively.

The remaining fine action duration $\hat{r}_{j^{*}}^{f}$ is estimated
similarly, but with the downward message as additional input:\vspace{-2mm}
\begin{align}
\tilde{h}_{j^{*}-1}^{f} & =\textrm{MLP}\left(\left[h_{j^{*}-1}^{f},x_{j^{*}}^{f},\tilde{a}_{j^{*}}^{f},\tilde{d}_{j^{*}}^{f},\tilde{m_{i}}^{c\rightarrow f}\right]\right);\qquad\hat{r}_{j^{*}}^{f}=\textrm{MLP}\left(\tilde{h}_{j^{*}-1}^{f}\right).
\end{align}
Effectively, the overall predicted duration of the interrupted action
is amended:\vspace{-2mm}
\begin{equation}
\hat{d}_{i^{*}}^{c}=\tilde{d}_{i^{*}}^{c}+\hat{r}_{i^{*}}^{c};\qquad\qquad\hat{d}_{j^{*}}^{f}=\tilde{d}_{j^{*}}^{f}+\hat{r}_{j^{*}}^{f}.
\end{equation}

After these refreshing steps, the hidden states $\tilde{h}_{i^{*}}^{c}$
and $\tilde{h}_{i^{*}}^{f}$ are passed to the $\Decoder$ as the
initial states, where the hierarchical prediction is rolled out further.

\paragraph{Model training.}

In $\Model$'s end-to-end training, we calculate the loss at each
level $l$ (among coarse and fine) and each stage $\star$ (among
the $\Encoder$, $\Transitioner$, and $\Decoder$) as a weighted
sum of negative log-likelihood loss (NLL) on predicted labels and
mean squared error (MSE) on predicted durations (for the $\Transitioner$
we only have the MSE loss):\vspace{-2mm}
\begin{align}
\mathcal{L}_{l}^{\star}= & \frac{1}{n_{l}}\sum_{k=1}^{n_{l}}\left[\lambda_{\textrm{label}}^{\star}\text{NLL}\left(\hat{x}_{k}^{l},x_{k}^{l}\right)+\lambda_{\textrm{duration}}^{\star}\text{MSE}\left(\hat{d}_{k}^{l},d_{k}^{l}\right)\right],\label{eq:encoder_or_decoder_loss}
\end{align}
and the total loss for $\Model$ is a sum of the losses in all layers
and stages:\vspace{-2mm}
\begin{align}
\mathcal{L}= & \sum_{l=1}^{L}\left[\mathcal{L}_{l}^{E}+\mathcal{L}_{l}^{R}+\mathcal{L}_{l}^{A}\right].\label{eq:total_loss}
\end{align}
The Encoder loss $\mathcal{L}_{l}^{E}$ is for regularizing the Encoder
and is optional. The weights $\lambda_{-}^{\star}$ are estimated
together with the network parameters by using a multi-task learning
framework similar to that of Kendall \etal \cite{kendall2018multi-task}.

For model validation, we use the videos of a single person from each
cross-validation split. We train $\Model$ for 20 epochs and select
the weights from the epoch with the lowest validation loss. We use
the ADAM \cite{kingma2014adam} optimizer with a learning rate of
$10^{-3}$ and a batch size of 512. The selected hidden size for the
GRUs and MLPs was 16. We used the PyTorch \cite{paszke2019pytorch}
framework for implementation of $\Model$.

\subsection{Data annotation\label{subsec:Breakfast-Actions-Annotation}}

To support the problem structure formulated above we reannotated the
Breakfast Actions videos \cite{kuehne2014language}, which is the
largest multi-level video activity dataset publicly available. This
dataset contains footage of 52 people preparing 10 distinct breakfast-related
dishes, totaling 1,717 videos. It originally contains fine- and coarse-level
annotations of the actions but the hierarchy is incoherent (inconsistent
semantic abstraction), incomplete (only 804 of the videos have fine-level
annotations), and statistically weak (many fine actions are less than
a few frames).

We employed two annotators working independently on all 1,717 videos
and one verifier who checked the consistency of the annotations. Following
the hierarchy definition in Sec.~\ref{subsec:Problem-Formulation},
we annotated a two-level hierarchy of \emph{coarse activities} and
\emph{fine actions}. Each label of activity or action follows the
format of \texttt{\small{}<verb-noun>} where verbs and nouns are selected
from a predefined vocabulary. The two vocabulary sets were built by
a pilot round of annotation. The coarse activities can share the fine
action labels. For instance, \texttt{\small{}<add-salt>} fine action
label can be used for many coarse activities including \texttt{\small{}<make-salad>},
\texttt{\small{}<fry-egg>}, and \texttt{\small{}<make-sandwich>}.
In actual annotation, we have 30 \texttt{\small{}<verb-noun>} pairs
for coarse activities and 140 for fine actions that are active. The
new annotation resulted in a total of 25,537 label-duration annotations
with 6,549 at the coarse level and 18,988 at the fine level. We call
the new annotation \emph{Hierarchical Breakfast} dataset and it is
available for download\footnote{https://github.com/RomeroBarata/hierarchical\_action\_prediction},
alongside the source code for $\Model$.

\subsection{Metrics\label{subsec:Metrics}}

Recent action prediction works \cite{farha2018when,ke2019time-conditioned}
widely used mean-over-class (MoC) as the key performance metric. However,
MoC is susceptible to bias in class imbalance which exists in action
prediction datasets. More importantly, as any frame-based metrics,
it merits any correctly predicted frames even when the predicted segments
are mostly unaligned due to under- or over-segmentation. We verified
these conceptual problems by setting up an experiment (detailed in
Sec.~\ref{sec:Experiments}) using an under-segmenting dummy predictor
that takes advantage of the flaw of the metric and win over state-of-the-art
methods on many settings. We call our dummy predictor ``under-segmenting''
because it predicts that the future consists simply of one single
long action.

In the search for better metrics, we examined options including the
segmental edit distance, the mean-over-frame (MoF), and the F1@$k$.
Among them, we learned that the most suitable metric for the purpose
of action prediction is the F1@$k$ for its robustness to variation
in video duration and minor shifts caused by annotation errors. Furthermore,
it penalizes both over- and under-segmentations such as from our dummy
predictor. This metric was previously used for temporal detection
and segmentation \cite{lea2017temporal}. Applied to the prediction
task, we first calculate the intersection over union (IoU) of the
predicted segments with the ground-truth. Any overlapping pair with
IoU surpassing the chosen threshold $0<k<1$ is counted as correct
when contributing to the final $\text{F}1=\nicefrac{2\times\textrm{Prec\ensuremath{\times}Recall}}{\text{Prec+Recall}}$.
\vspace{-2mm}

\section{Experiments\label{sec:Experiments}}

\subsection{Experiment settings}

We setup experiments following the common settings in which 20\% or
30\% of the videos are observed and the prediction is done on the
remaining portion (70\% or 80\%). We also follow previous convention
to use annotated labels of the observed portion to be the input, with
the assumption that in practical applications these labels can be
reliably provided by action detection engines \cite{carreira2017quo}.
All experiments are done with 4-fold cross--validation as in previous
works \cite{farha2018when,ke2019time-conditioned}.

We use the new \emph{Hierarchical Breakfast} (see Sec.~\ref{subsec:Breakfast-Actions-Annotation})
as our principal source of data for its most comprehensive multi-level
activities. Besides this one, the 50 Salads dataset \cite{stein2013combining}
also has two-level annotations and was used in several previous works
\cite{farha2018when,ke2019time-conditioned}. However per acquisition
procedure description \cite{stein2013combining} and through our independent
examination, we concluded that 50 Salads is only suitable for action
detection and segmentation but not for action prediction because of
the randomness in scripted action sequences. Such scripts were generated
from an artificial statistical model that intentionally introduces
random sequences instead of following natural human behaviors. This
makes most non-trivial methods converge to similar prediction results
(reported in the supplementary material), and hence is not suitable
for differentiating their performances.

\subsection{Metrics and data assessment}

We set up experiments to demonstrate the drawback of MoC and verify
the robustness of F1@$k$ described in Sec.~\ref{subsec:Metrics}.
We use a dummy predictor that simply predicts that the interrupted
action goes on for the rest of the video, i.e., the extreme under-segmentation.
We compare this dummy predictor to the results reported by two top
performing predictors by Farha \etal \cite{farha2018when} and
Ke \etal \cite{ke2019time-conditioned}\footnote{We could neither obtain nor reproduce the implementation of Ke \etal
\cite{ke2019time-conditioned}, therefore we could only use the reported
performance on the original Breakfast annotation and MoC metrics (last
row of Table \ref{tab:moc-breakfast-coarse-original}).} at the coarse-level of the original Breakfast Actions dataset. As
results in Table~\ref{tab:moc-breakfast-coarse-original} show, the
dummy predictor performs comparable to the best and usually outperforms
one or both of the methods in MoC by exploiting its fragility toward
over- and under- segmentation.

\begin{table}
\caption{Mean-over-class (MoC) scores on the coarse-level of the original Breakfast
Actions dataset. The dummy predictor matches performance with state-of-the-art
methods, which demonstrates the weakness of the MoC metric.\label{tab:moc-breakfast-coarse-original}}
\resizebox{\textwidth}{!}{%
\centering{}%
\begin{tabular}{lcccc|cccc}
\hline 
Observe & \multicolumn{4}{c|}{20\%} & \multicolumn{4}{c}{30\%}\tabularnewline
\cline{2-9} \cline{3-9} \cline{4-9} \cline{5-9} \cline{6-9} \cline{7-9} \cline{8-9} \cline{9-9} 
\noalign{\vskip0.1cm}
Predict & 10\% & 20\% & 30\% & 50\% & 10\% & 20\% & 30\% & 50\%\tabularnewline
\hline 
\noalign{\vskip0.1cm}
Dummy & 0.64 & 0.51 & 0.44 & 0.35 & 0.68 & 0.54 & 0.44 & 0.36\tabularnewline
\noalign{\vskip0.1cm}
Farha \etal \cite{farha2018when} & 0.60 & 0.50 & 0.45 & 0.40 & 0.61 & 0.50 & 0.45 & 0.42\tabularnewline
\noalign{\vskip0.1cm}
Ke \etal \cite{ke2019time-conditioned} & 0.64 & 0.56 & 0.50 & 0.44 & 0.66 & 0.56 & 0.49 & 0.44\tabularnewline
\hline 
\noalign{\vskip0.1cm}
\end{tabular}}\vspace{-2mm}
\end{table}

When we replace MoC with our chosen F1@$k$ metric (Table~\ref{tab:f1-breakfast-coarse-original}),
the dummy predictor only has good scores at the immediate 10\% prediction
(as designed) and marked down significantly afterward as continuing
action no longer matches with the actual events.

\begin{table}
\caption{F1@\textbf{0.25} scores on the coarse-level of the original Breakfast
Actions dataset. The F1@\textbf{0.25} metric is robust to the dummy
predictor and helps better methods to stand out in long-term predictions.\label{tab:f1-breakfast-coarse-original}}
\resizebox{\textwidth}{!}{%
\centering{}%
\begin{tabular}{lcccc|cccc}
\hline 
Observe & \multicolumn{4}{c|}{20\%} & \multicolumn{4}{c}{30\%}\tabularnewline
\cline{2-9} \cline{3-9} \cline{4-9} \cline{5-9} \cline{6-9} \cline{7-9} \cline{8-9} \cline{9-9} 
\noalign{\vskip0.1cm}
Predict & 10\% & 20\% & 30\% & 50\% & 10\% & 20\% & 30\% & 50\%\tabularnewline
\hline 
\noalign{\vskip0.1cm}
Dummy & 0.77 & 0.61 & 0.51 & 0.34 & 0.80 & 0.67 & 0.56 & 0.40\tabularnewline
\noalign{\vskip0.1cm}
Farha \etal \cite{farha2018when} & 0.76 & 0.68 & 0.64 & 0.58 & 0.78 & 0.71 & 0.68 & 0.64\tabularnewline
\hline 
\noalign{\vskip0.1cm}
\end{tabular}}\vspace{-2mm}
\end{table}

\subsection{Predicting activity hierarchy}

In this section, we validate the performance of our proposed $\Model$
against reference methods. As predicting the roll-out of activity
hierarchy is a new task, we implemented several baselines and adapted
a state-of-the-art method by Farha \etal \cite{farha2018when}
to work with two-level activities. 

All baselines accept observed labels and (accumulated) durations and
output those of future events. The first baseline, \emph{Independent-Single-RNN,}
uses two separate GRUs for coarse activities and fine actions hence
does not consider the correlation between the two levels. To take
into account this correlation, \emph{Joint-Single-RNN}, the second
baseline, models the joint distribution of the two processes by concatenating
input from both levels and predicting them together. The third baseline,
\emph{Synced-Pair-RNN}, is more sophisticated and has two parallel
GRUs for the two levels operating at the same clock, which communicate
regularly by one-way coarse-to-fine messages. Because the last two
baselines operate with a single recurrent clock on two signals that
are not synchronized, coarse level inputs are repeated as needed to
sync-up with the fine level counterparts.

The original Farha \etal's model \cite{farha2018when} (denoted
as ``Farha'') only accepts a single level of observed action as
input, hence two separated instances of it are used to predict at
coarse and fine level. To make consistent competition, we extend this
method to accept hierarchical input by jointly observing and predicting
the two levels of actions (named ``Farha2'').

\begin{figure}
\centering{}%
\begin{tabular}{c|c}
\includegraphics[height=0.275\textwidth]{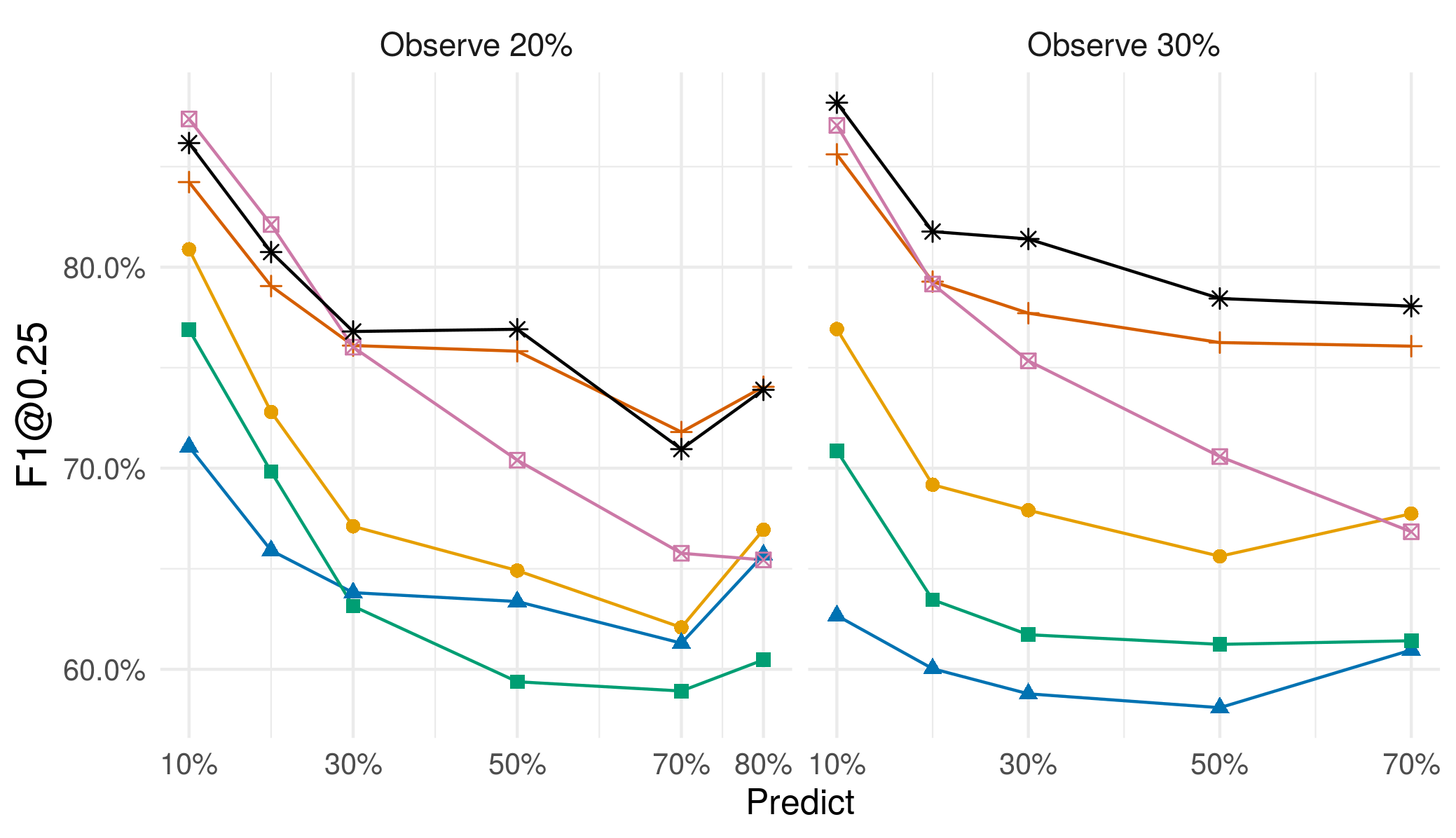} & \includegraphics[height=0.275\textwidth]{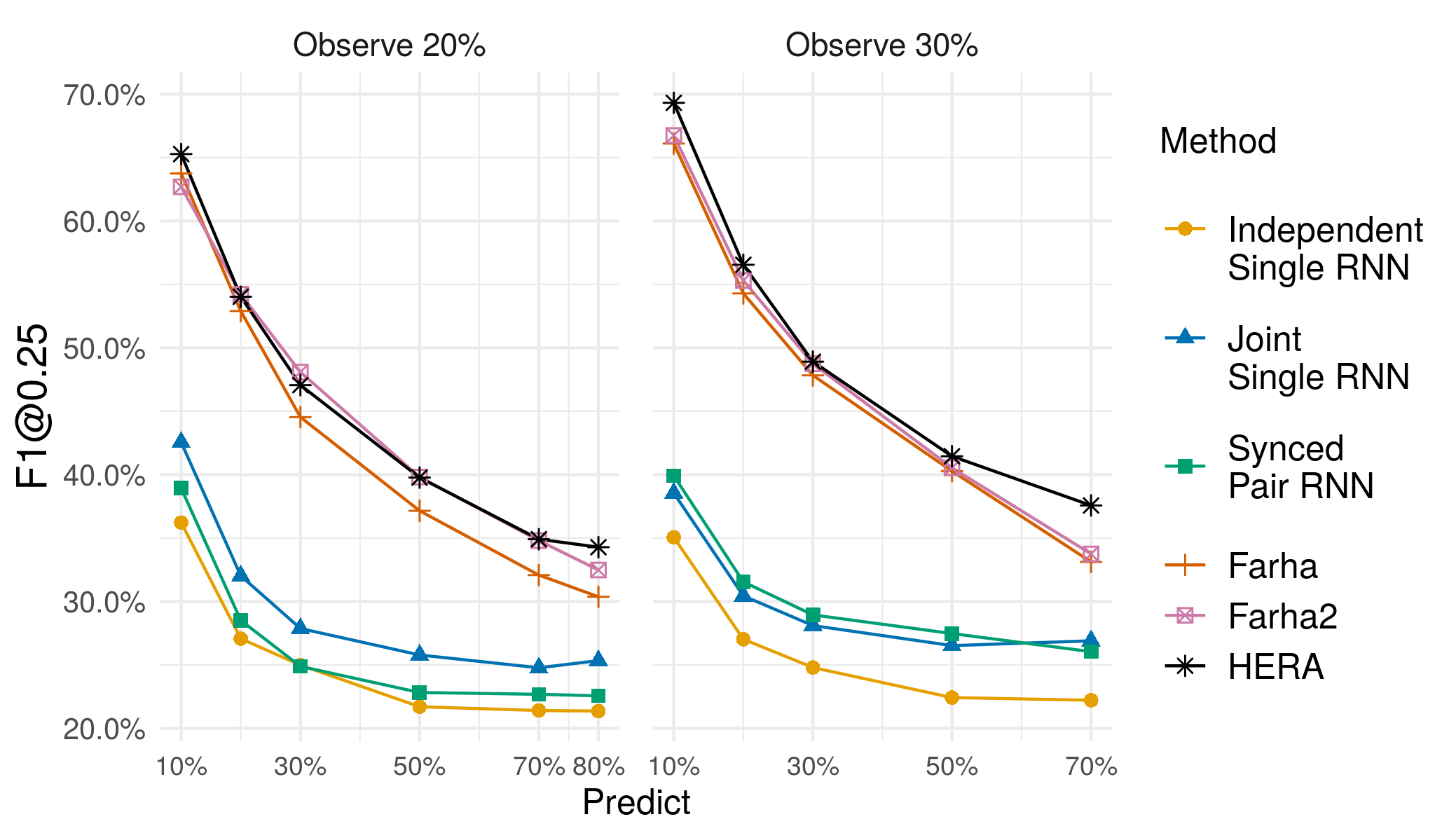}\tabularnewline
\multicolumn{1}{c}{Coarse Activities} & Fine Actions\tabularnewline
\end{tabular}\medskip{}
\caption{F1@\textbf{0.25} performance of $\protect\Model$ and related baselines
on \textbf{coarse} (left fig.) and \textbf{fine} (right fig.) levels
of Hierarchical Breakfast dataset.}
\label{fig:breakfast_f1at25}\vspace{-2mm}
\end{figure}

\begin{figure}
\centering{}\includegraphics[width=0.9\textwidth]{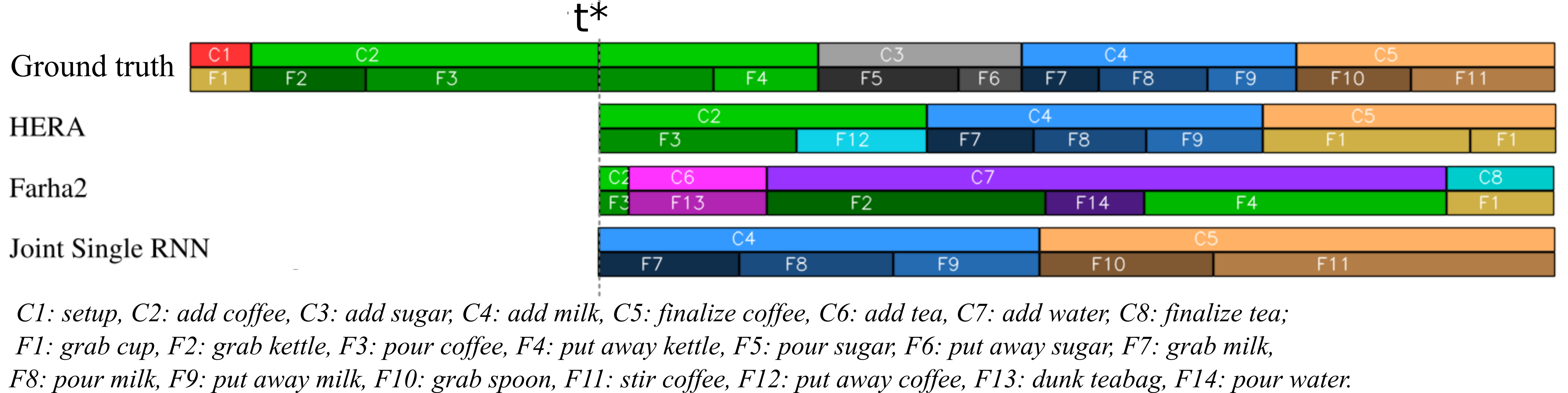}\medskip{}
\caption{Qualitative evaluation of predictions of different methods on task
``make coffee''. The first timeline shows the observed and ground-truth
future. Others show future predictions of corresponding methods.}
\label{fig:qualitative-hera}\vspace{-2mm}
\end{figure}

We compare the performance of HERA and aforementioned reference models
on Hierarchical Breakfast. The F1@0.25 scores on two settings are
shown in Fig.~\ref{fig:breakfast_f1at25}. The performance results
suggest notable patterns. First of all, we learned that coarse activities
and fine actions have strong but asymmetrical correlation. When modeled
in a joint distribution, the fine action prediction (right subfigure)
is improved over independent modeling (Joint-Single-RNN to Independent-Single-RNN,
and Farha 2 to Farha); meanwhile coarse prediction (left subfigure)
is degraded by the noise introduced in the over-detailed data from
the fine channel. 

Secondly, modeling coarse and fine as parallel interactive concurrent
processes (Synced-Pair-RNN) may help especially in encoding long observation.
However the na\"ive synchronicity between the two processes is unreliable
and in many cases significantly hurt the performance. Thirdly, when
introducing structure to the cross-process interaction (HERA), the
performance significantly improved both near- and far- term and across
coarse- and fine- channels. This result suggests that appropriate
structured modeling is key to deeply observe, understand and generate
hierarchical activity patterns. Fourthly, in longer term prediction,
asynchronously clocked RNNs (as in HERA) alleviate the error accumulation
issues persistent in all other synchronous RNN based models. 

Overall, $\Model$ attained higher prediction accuracy in relation
to other methods, especially in harder cases such as on far-term fine
actions. To further understand this improvement, we visualize the
predictions of $\Model$ and compare them with those of other methods.
One example is shown in Fig.~\ref{fig:qualitative-hera} and more
are included in the supplementary material. In this case, $\Model$
predicts most accurately the remaining duration of the unfinished
actions thanks to the special design of the $\Transitioner$. Furthermore,
the strong coarse-to-fine structure helps it recover from mistakes
while maintaining cross-hierarchy consistency. Without this structure,
other methods tend to roll-out on top of the mistakes and cannot recover.
They also sometimes allow ill-formed hierarchy such as the pink \texttt{\small{}C6-F13
}family in Farha2.

We argued that the MoC metric is not appropriate for the problem at
hand, but we report it next for transparency and completeness. For
observing 20\% and predicting the next 10\%/20\%/30\%/50\%/70\%/80\%,
$\Model$ attained an MoC of 0.77/0.68/0.57/0.51/0.51/0.57 for the
coarse level, and 0.42/0.31/0.26/0.23/0.21 /0.21 for the fine level;
Farha2 attained an MoC of 0.76/0.67/0.57/0.51/0.50 /0.52 for the coarse
level, and an MoC of 0.39/0.29/0.26/0.22/0.21/0.20 for the fine level.
For observing $30\%$, and predicting the next 10\%/20\%/30\%/50\%/70\%,
$\Model$ attained an MoC of 0.77/0.62/0.59/0.53/0.61 for the coarse
level, and an MoC of 0.44/0.33/0.28/0.25/0.23 for the fine level,
whereas Farha2 attained an MoC of 0.76/0.64/0.58/0.52/0.55 for the
coarse level, and an MoC of 0.41/0.32 /0.28/0.24/0.22 for the fine
level. $\Model$'s MoC is higher than Farha2's MoC in most cases,
but as discussed earlier the F1@k metric should be preferred when
comparing models for action prediction.

\paragraph{Ablation study.}

To further understand the roles of components and design choices in
$\Model$, we switch off several key aspects of the model and observe
the change in performance (Table~\ref{tab:new-breakfast-f1@0.25-ablation}).
The first variation without the two-way cross-level messaging suffers
significant performance loss as the correlation between the two channels
is ignored. The second variation lacks the explicit coarse label $\hat{x}_{i+1}^{c}$
in downward messages and slightly under performs as missing the direct
strong ``instruction'' of the discrete coarse labels. Lastly, the
third row provides evidence for the importance of the $\Transitioner$
stage in wrapping up unfinished action consistently at all levels.

\begin{table}
\centering{}\caption{F1@\textbf{0.25} scores of $\protect\Model$ and its variations on
the fine-level of Hierarchical Breakfast dataset with 20\% of the
videos observed.\label{tab:new-breakfast-f1@0.25-ablation}}
\begin{tabular}{lccccc}
\toprule 
Variations & 10\% & 20\% & 30\% & 50\% & 80\%\tabularnewline
\midrule
W/o $\downarrow\uparrow$ msg & 0.36 & 0.27 & 0.25 & 0.22 & 0.21\tabularnewline
W/o label in $\downarrow$ msg & 0.62 & 0.51 & 0.45 & 0.36 & 0.31\tabularnewline
W/ dis. $\Transitioner$ & 0.64 & 0.53 & 0.45 & 0.37 & \textbf{0.34}\tabularnewline
\midrule 
Full HERA & \textbf{0.65} & \textbf{0.54} & \textbf{0.47} & \textbf{0.40} & \textbf{0.34}\tabularnewline
\bottomrule
\end{tabular}\vspace{-2mm}
\end{table}

\vspace{-2mm}

\section{Conclusions}

We have introduced $\Model$ (Hierarchical Encoder-Refresher-Anticipator),
a new hierarchical neural network for modeling and predicting the
long-term multi-level action dynamics in videos. To promote further
research we re-annotated from scratch 1,717 videos in the Breakfast
Actions dataset, creating a new and complete semantically coherent
annotation of activity hierarchy, which we named \emph{Hierarchical
Breakfast}. We also reassessed the commonly used MoC metric in action
prediction, and found it unreliable for the task. As a result we investigated
multiple metrics and found the F1@$k$ metric to reflect human activity
best among them. We demonstrated that our $\Model$ naturally handles
hierarchically structured activities, including interruptions in the
observed activity hierarchy. When compared to related methods that
do not exploit the hierarchical structure in human activities, or
explore it in a sub-optimal way, $\Model$ attained superior results
specially in the long-term regime.

\vspace{-2mm}

\section*{Acknowledgments}

We would like to thank Thao Minh Le for the helpful discussions in
regards to building the \emph{Hierarchical Breakfast} dataset.

\bibliography{arxiv}

\begin{thebibliography}{10}

\bibitem{farha2019uncertainty-aware}
Yazan Abu~Farha and Juergen Gall.
\newblock {Uncertainty-Aware} anticipation of activities.
\newblock In {\em 2019 {IEEE/CVF} International Conference on Computer Vision
  Workshop ({ICCVW})}, pages 1197--1204, October 2019.

\bibitem{ajzen1985intentions}
Icek Ajzen.
\newblock From intentions to actions: A theory of planned behavior.
\newblock In {\em Action Control: From Cognition to Behavior}, pages 11--39.
  Springer Berlin Heidelberg, 1985.

\bibitem{carreira2017quo}
Joao Carreira and Andrew Zisserman.
\newblock Quo vadis, action recognition? a new model and the kinetics dataset.
\newblock In {\em 2017 {IEEE} Conference on Computer Vision and Pattern
  Recognition ({CVPR})}, pages 4724--4733, July 2017.

\bibitem{cho2014learning}
Kyunghyun Cho, Bart van Merrienboer, Caglar Gulcehre, Dzmitry Bahdanau, Fethi
  Bougares, Holger Schwenk, and Yoshua Bengio.
\newblock Learning phrase representations using {RNN} {Encoder--Decoder} for
  statistical machine translation.
\newblock In {\em Proceedings of the 2014 Conference on Empirical Methods in
  Natural Language Processing ({EMNLP})}, pages 1724--1734. ACL, 2014.

\bibitem{chung2016hierarchical}
Junyoung Chung, Sungjin Ahn, and Yoshua Bengio.
\newblock Hierarchical multiscale recurrent neural networks.
\newblock In {\em 5th International Conference on Learning Representations,
  {ICLR} 2017}. OpenReview.net, April 2017.

\bibitem{duong2009efficient}
Thi Duong, Dinh Phung, Hung Bui, and Svetha Venkatesh.
\newblock Efficient duration and hierarchical modeling for human activity
  recognition.
\newblock {\em Artificial intelligence}, 173(7):830--856, May 2009.

\bibitem{farha2018when}
Yazan~Abu Farha, Alexander Richard, and Juergen Gall.
\newblock When will you do what? - anticipating temporal occurrences of
  activities.
\newblock In {\em 2018 {IEEE/CVF} Conference on Computer Vision and Pattern
  Recognition}, pages 5343--5352. IEEE, June 2018.

\bibitem{gammulle2019forecasting}
Harshala Gammulle, Simon Denman, Sridha Sridharan, and Clinton Fookes.
\newblock Forecasting future action sequences with neural memory networks.
\newblock In {\em Proceedings of the British Machine Vision Conference 2019}.
  British Machine Vision Association, September 2019.

\bibitem{hihi1996hierarchical}
Salah~El Hihi and Yoshua Bengio.
\newblock Hierarchical recurrent neural networks for {Long-Term} dependencies.
\newblock In {\em Advances in Neural Information Processing Systems 8}, pages
  493--499. MIT Press, 1996.

\bibitem{ke2019time-conditioned}
Qiuhong Ke, Mario Fritz, and Bernt Schiele.
\newblock {Time-Conditioned} action anticipation in one shot.
\newblock In {\em 2019 {IEEE/CVF} Conference on Computer Vision and Pattern
  Recognition ({CVPR})}, pages 9917--9926. IEEE, June 2019.

\bibitem{kendall2018multi-task}
Alex Kendall, Yarin Gal, and Roberto Cipolla.
\newblock Multi-task learning using uncertainty to weigh losses for scene
  geometry and semantics.
\newblock In {\em 2018 {IEEE/CVF} Conference on Computer Vision and Pattern
  Recognition}, pages 7482--7491, June 2018.

\bibitem{kingma2014adam}
Diederik~P Kingma and Jimmy Ba.
\newblock Adam: A method for stochastic optimization.
\newblock In {\em 3rd International Conference on Learning Representations,
  {ICLR} 2015}, 2015.

\bibitem{koutnik2014clockwork}
Jan Koutnik, Klaus Greff, Faustino Gomez, and Juergen Schmidhuber.
\newblock A clockwork {RNN}.
\newblock In Eric~P Xing and Tony Jebara, editors, {\em Proceedings of the 31st
  International Conference on Machine Learning}, volume~32 of {\em Proceedings
  of Machine Learning Research}, pages 1863--1871, Bejing, China, 2014. PMLR.

\bibitem{kuehne2014language}
Hilde Kuehne, Ali Arslan, and Thomas Serre.
\newblock The language of actions: Recovering the syntax and semantics of
  {Goal-Directed} human activities.
\newblock In {\em 2014 {IEEE} Conference on Computer Vision and Pattern
  Recognition}, pages 780--787, June 2014.

\bibitem{lan2014hierarchical}
Tian Lan, Tsung-Chuan Chen, and Silvio Savarese.
\newblock A hierarchical representation for future action prediction.
\newblock In {\em Computer Vision -- {ECCV} 2014}, pages 689--704. Springer
  International Publishing, 2014.

\bibitem{lea2017temporal}
Colin Lea, Michael~D Flynn, Ren{\'{e}} Vidal, Austin Reiter, and Gregory~D
  Hager.
\newblock Temporal convolutional networks for action segmentation and
  detection.
\newblock In {\em 2017 {IEEE} Conference on Computer Vision and Pattern
  Recognition ({CVPR})}, pages 1003--1012, July 2017.

\bibitem{mahmud2017joint}
Tahmida Mahmud, Mahmudul Hasan, and Amit~K Roy-Chowdhury.
\newblock Joint prediction of activity labels and starting times in untrimmed
  videos.
\newblock In {\em 2017 {IEEE} International Conference on Computer Vision
  ({ICCV})}, pages 5784--5793, October 2017.

\bibitem{mehrasa2019variational}
Nazanin Mehrasa, Akash~Abdu Jyothi, Thibaut Durand, Jiawei He, Leonid Sigal,
  and Greg Mori.
\newblock A variational {Auto-Encoder} model for stochastic point processes.
\newblock In {\em 2019 {IEEE/CVF} Conference on Computer Vision and Pattern
  Recognition ({CVPR})}, pages 3160--3169. IEEE, June 2019.

\bibitem{miech2019leveraging}
Antoine Miech, Ivan Laptev, Josef Sivic, Heng Wang, Lorenzo Torresani, and
  Du~Tran.
\newblock Leveraging the present to anticipate the future in videos.
\newblock In {\em 2019 {IEEE/CVF} Conference on Computer Vision and Pattern
  Recognition Workshops ({CVPRW})}, pages 2915--2922. IEEE, June 2019.

\bibitem{paszke2019pytorch}
Adam Paszke, Sam Gross, Francisco Massa, Adam Lerer, James Bradbury, Gregory
  Chanan, Trevor Killeen, Zeming Lin, Natalia Gimelshein, Luca Antiga, Alban
  Desmaison, Andreas Kopf, Edward Yang, Zachary DeVito, Martin Raison, Alykhan
  Tejani, Sasank Chilamkurthy, Benoit Steiner, Lu~Fang, Junjie Bai, and Soumith
  Chintala.
\newblock {PyTorch}: An imperative style, {High-Performance} deep learning
  library.
\newblock In {\em Advances in Neural Information Processing Systems 32}, pages
  8026--8037. 2019.

\bibitem{qi2017predicting}
Siyuan Qi, Siyuan Huang, Ping Wei, and Song{-}Chun Zhu.
\newblock Predicting human activities using stochastic grammar.
\newblock In {\em 2017 {IEEE} International Conference on Computer Vision
  ({ICCV})}, pages 1173--1181, October 2017.

\bibitem{stein2013combining}
Sebastian Stein and Stephen~J McKenna.
\newblock Combining embedded accelerometers with computer vision for
  recognizing food preparation activities.
\newblock In {\em Proceedings of the 2013 {ACM} international joint conference
  on Pervasive and ubiquitous computing}, pages 729--738. ACM, September 2013.

\bibitem{Truyen:2006}
Truyen~T Tran, Dinh~Q Phung, Svetha Venkatesh, and Hung~H Bui.
\newblock {AdaBoost.MRF}: Boosted markov random forests and application to
  multilevel activity recognition.
\newblock In {\em 2006 {IEEE} Computer Society Conference on Computer Vision
  and Pattern Recognition ({CVPR'06})}, volume~2, pages 1686--1693, June 2006.

\bibitem{yan2018hierarchical}
Shiyang Yan, Jeremy~S Smith, Wenjin Lu, and Bailing Zhang.
\newblock Hierarchical multi-scale attention networks for action recognition.
\newblock {\em Signal Processing: Image Communication}, 61:73--84, February
  2018.

\bibitem{zhong2018time}
Yatao Zhong, Bicheng Xu, Guang-Tong Zhou, Luke Bornn, and Greg Mori.
\newblock Time perception machine: Temporal point processes for the when, where
  and what of activity prediction.
\newblock {\em arXiv preprint arXiv:1808.04063}, 2018.

\end{thebibliography}
\bibliographystyle{plain}

\appendix

\section*{Supplementary Material}

\section{Hierarchical Breakfast Annotation Analysis}

We annotated 1717 videos into a two-level hierarchy: coarse activities
and fine actions. This resulted in 25537 annotated segments, with
6549 of them being coarse activities and 18988 of them being fine
actions. At the end of the annotation, there were 30 unique coarse
activities and 140 unique fine actions annotated across the whole
dataset.

\begin{figure}
\centering{}\includegraphics[width=1\textwidth]{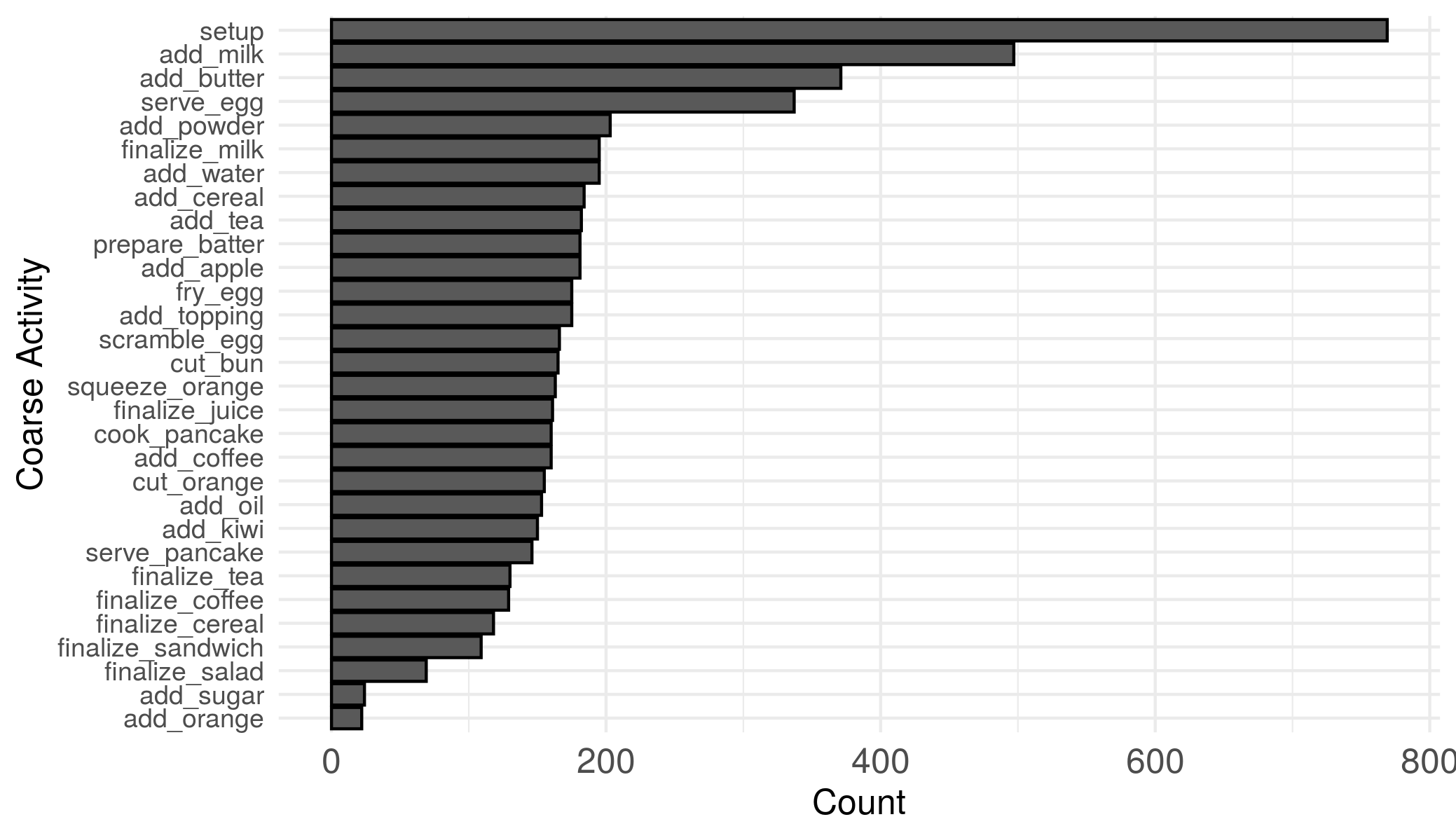}\caption{Distribution of the annotated coarse activities for the Hierarchical
Breakfast Actions dataset. Coarse activity name is shown on the y-axis
whereas the number of times the activity appeared is shown on the
x-axis.}
\label{fig:coarse-activity-dist}
\end{figure}

\begin{figure}
\centering{}\includegraphics[width=1\textwidth]{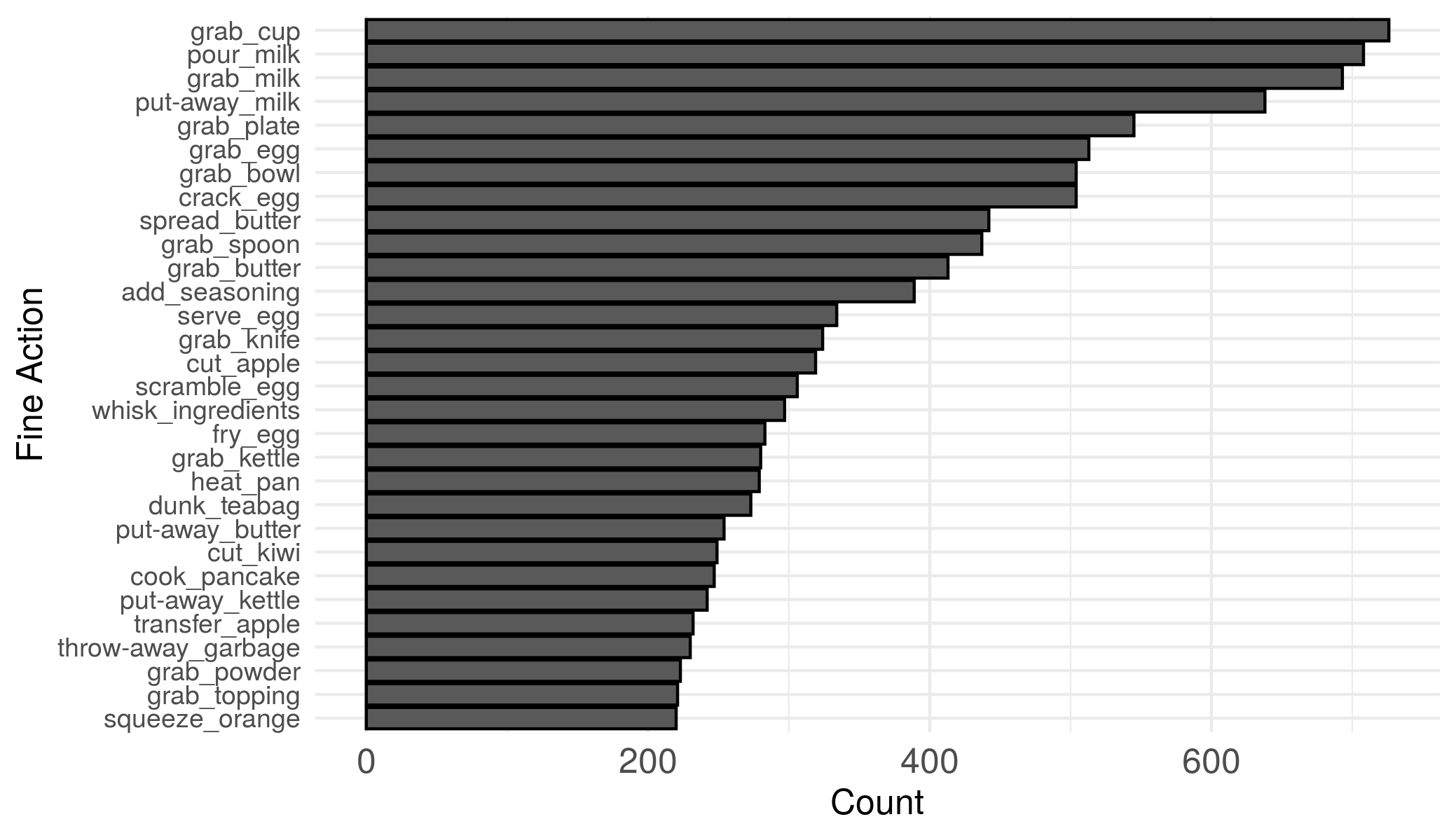}\caption{Distribution of the top-30 annotated fine actions for the Hierarchical
Breakfast Actions dataset. We show here only the top-30 fine actions
to avoid clutter. Fine action name is shown on the y-axis whereas
the number of times the action appeared is shown on the x-axis.}
\label{fig:fine-action-dist}
\end{figure}

In Fig.~\ref{fig:coarse-activity-dist} we can see the number of
times each coarse activity got annotated. In Fig.~\ref{fig:fine-action-dist}
we can see the number of times the top 30 fine actions were annotated
(we show the top 30 to avoid clutter). Some activities are not frequent,
since the preparation of breakfast meals can widely vary from person
to person. For instance, not everyone add sugar to their coffee. These
variations in behavior are natural and were all annotated.

\section{Additional Results}

\subsection{Hierarchical Breakfast Dataset}

The F1@\textbf{0.25} values for the results in Fig.~3 of the main
paper are shown here on Table \ref{tab:new-breakfast-f1@0.25}.

\begin{table*}
\centering{}\caption{F1@\textbf{0.25} of HERA and related methods on the Hierarchical Breakfast
Actions dataset. For this experiment, the methods are allowed to observe
a percentage of the video (20\% or 30\%) and need to predict the whole
unseen future (70\% or 80\%). The results are an average of a 4-fold
cross-validation and higher results are better.\label{tab:new-breakfast-f1@0.25}}
\resizebox{\textwidth}{!}{
\begin{tabular}{cccccccccccccc}
\toprule 
 & Observe & \multicolumn{6}{c}{20\%} &  & \multicolumn{5}{c}{30\%}\tabularnewline
\cmidrule{3-8} \cmidrule{4-8} \cmidrule{5-8} \cmidrule{6-8} \cmidrule{7-8} \cmidrule{8-8} \cmidrule{10-14} \cmidrule{11-14} \cmidrule{12-14} \cmidrule{13-14} \cmidrule{14-14} 
 & Predict & 10\% & 20\% & 30\% & 50\% & 70\% & 80\% &  & 10\% & 20\% & 30\% & 50\% & 70\%\tabularnewline
\midrule 
\multirow{6}{*}{Coarse} & Dummy & 87.3\% & 76.9\% & 68.1\% & 53.7\% & 42.6\% & 36.4\% &  & 88.0\% & 77.5\% & 69.9\% & 55.4\% & 40.1\%\tabularnewline
 & Baseline 0 & 80.9\% & 72.8\% & 67.1\% & 64.9\% & 62.1\% & 66.9\% &  & 76.9\% & 69.2\% & 67.9\% & 65.6\% & 67.7\%\tabularnewline
 & Baseline 1 & 71.1\% & 65.9\% & 63.8\% & 63.4\% & 61.3\% & 65.7\% &  & 62.7\% & 60.0\% & 58.8\% & 58.1\% & 61.0\%\tabularnewline
 & Baseline 2 & 76.9\% & 69.8\% & 63.1\% & 59.4\% & 58.9\% & 60.5\% &  & 70.9\% & 63.5\% & 61.7\% & 61.2\% & 61.4\%\tabularnewline
 & Farha \etal \cite{farha2018when} & 84.2\% & 79.1\% & 76.1\% & 75.8\% & \textbf{71.8\%} & 74.0\% &  & 85.6\% & 79.3\% & 77.7\% & 76.2\% & 76.1\%\tabularnewline
 & Farha2 \etal \cite{farha2018when} & \textbf{87.4\%} & \textbf{82.1\%} & 76.0\% & 70.4\% & 65.8\% & 65.4\% &  & 87.0\% & 79.2\% & 75.3\% & 70.6\% & 66.8\%\tabularnewline
\midrule 
\multirow{6}{*}{Fine} & Dummy & 62.2\% & 41.8\% & 29.6\% & 16.7\% & 10.3\% & 7.5\% &  & 66.0\% & 46.4\% & 35.4\% & 21.8\% & 12.1\%\tabularnewline
 & Baseline 0 & 36.2\% & 27.1\% & 25.0\% & 21.7\% & 21.4\% & 21.4\% &  & 35.1\% & 27.0\% & 24.8\% & 22.4\% & 22.2\%\tabularnewline
 & Baseline 1 & 42.5\% & 32.0\% & 27.9\% & 25.8\% & 24.8\% & 25.3\% &  & 38.5\% & 30.4\% & 28.1\% & 26.5\% & 26.9\%\tabularnewline
 & Baseline 2 & 38.9\% & 28.5\% & 24.9\% & 22.8\% & 22.7\% & 22.6\% &  & 39.9\% & 31.5\% & 28.9\% & 27.5\% & 26.0\%\tabularnewline
 & Farha \etal \cite{farha2018when} & 63.7\% & 52.9\% & 44.5\% & 37.1\% & 32.1\% & 30.4\% &  & 66.1\% & 54.3\% & 47.8\% & 40.3\% & 33.1\%\tabularnewline
 & Farha2 \etal \cite{farha2018when} & 62.7\% & \textbf{54.2\%} & \textbf{48.1\%} & \textbf{39.8\%} & 34.8\% & 32.5\% &  & 66.7\% & 55.3\% & 48.7\% & 40.5\% & 33.7\%\tabularnewline
\midrule 
Coarse & \multirow{2}{*}{HERA} & 86.2\% & 80.7\% & \textbf{76.8\%} & \textbf{76.9\%} & 70.9\% & 73.9\% &  & \textbf{88.2\%} & \textbf{81.8\%} & \textbf{81.4\%} & \textbf{78.4\%} & \textbf{78.1\%}\tabularnewline
Fine &  & \textbf{65.3\%} & 54.0\% & 47.1\% & \textbf{39.8\%} & \textbf{34.9\%} & \textbf{34.3\%} &  & \textbf{69.3\%} & \textbf{56.5\%} & \textbf{48.9\%} & \textbf{41.5\%} & \textbf{37.6\%}\tabularnewline
\bottomrule
\end{tabular}}
\end{table*}

\begin{figure}
\begin{centering}
\includegraphics[width=1\textwidth]{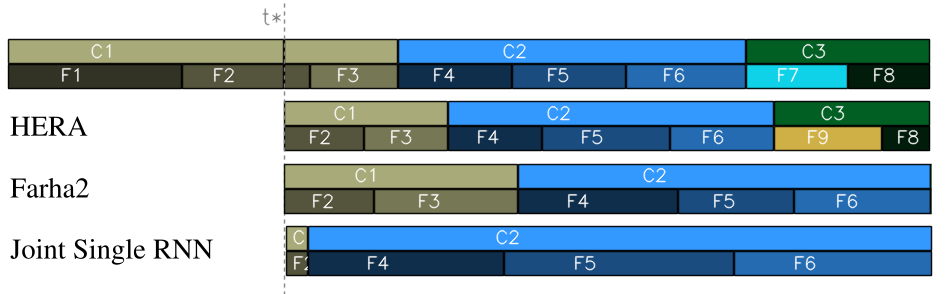}\caption{Qualitative evaluation of predictions of different methods on task
``prepare cereal''. The first timeline shows the observed and ground-truth
future. Others show future predictions of corresponding methods. C1:
add cereal, C2: add milk, C3: finalize cereal; F1: grab cereal, F2:
pour cereal, F3: put away cereal, F4: grab milk, F5: pour milk, F6:
put away milk, F7: stir cereal, F8: grab bowl, F9: grab spoon.}
\label{fig:qualitative-cereal}
\par\end{centering}
\end{figure}

\begin{figure}

\begin{centering}
\includegraphics[width=1\textwidth]{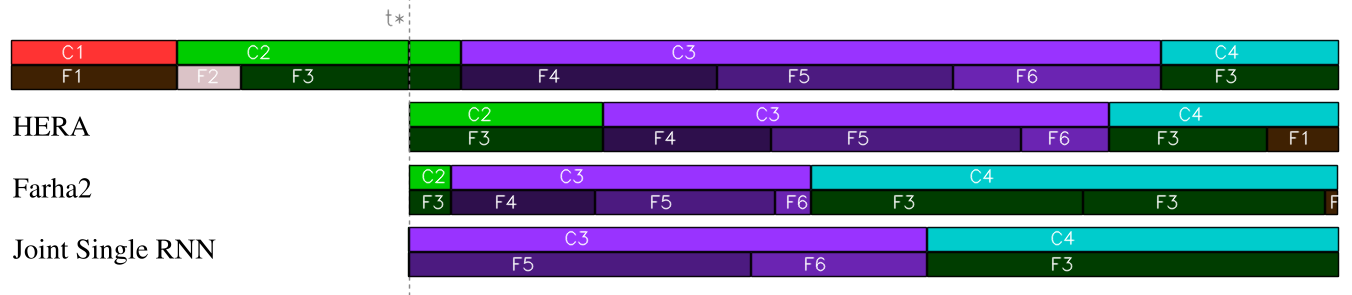}\caption{Qualitative evaluation of predictions of different methods on task
``prepare cereal''. The first timeline shows the observed and ground-truth
future. Others show future predictions of corresponding methods. C1:
setup, C2: add tea, C3: add water, C4: finalize tea; F1: grab cup,
F2: grab teabag, F3: dunk teabag, F4: grab kettle, F5: pour water,
F6: put away kettle.}
\label{fig:qualitative-tea}
\par\end{centering}
\end{figure}

Additional qualitative results are shown in Figs.~\ref{fig:qualitative-cereal}
and \ref{fig:qualitative-tea}. In these two examples, we can see
that in the short-term both HERA and Farha2 make predictions well
aligned with the ground-truth (e.g. F2 and F3 in Fig.~\ref{fig:qualitative-cereal}),
but as we move towards long-term predictions mistakes made by Farha2
in the fine-level quickly accumulate and generate misaligned predictions.
In Fig.~\ref{fig:qualitative-cereal}, for instance, F4 was too long
and from this point on Farha2 predictions F5 and F6 completely misaligned
with the ground-truth. HERA, on the other hand had more success in
correctly aligning the predicted fine actions with the ground-truth
since these predictions built on successful predictions at the coarse
level.

\subsection{50 Salads Dataset}

The F1@\textbf{0.25} attained by HERA and related methods are shown
on Table \ref{tab:50salads-f1@0.25}.

\begin{table*}
\centering{}\caption{F1@\textbf{0.25} of HERA and related methods on the mid and fine levels
of the 50 Salads dataset. For this experiment, the methods are allowed
to observe a percentage of the video (20\% or 30\%) and need to predict
the whole unseen future (70\% or 80\%). The results are an average
of a 5-fold cross-validation and higher results are better.\label{tab:50salads-f1@0.25}}
\resizebox{\textwidth}{!}{
\begin{tabular}{cccccccccccccc}
\toprule 
 & Observe & \multicolumn{6}{c}{20\%} &  & \multicolumn{5}{c}{30\%}\tabularnewline
\cmidrule{3-8} \cmidrule{4-8} \cmidrule{5-8} \cmidrule{6-8} \cmidrule{7-8} \cmidrule{8-8} \cmidrule{10-14} \cmidrule{11-14} \cmidrule{12-14} \cmidrule{13-14} \cmidrule{14-14} 
 & Predict & 10\% & 20\% & 30\% & 50\% & 70\% & 80\% &  & 10\% & 20\% & 30\% & 50\% & 70\%\tabularnewline
\cmidrule{1-8} \cmidrule{2-8} \cmidrule{3-8} \cmidrule{4-8} \cmidrule{5-8} \cmidrule{6-8} \cmidrule{7-8} \cmidrule{8-8} \cmidrule{10-14} \cmidrule{11-14} \cmidrule{12-14} \cmidrule{13-14} \cmidrule{14-14} 
\multirow{5}{*}{Mid} & Dummy & 46.2\% & 23.5\% & 12.3\% & 1.1\% & 0.0\% & 0.0\% &  & \textbf{49.1\%} & 23.3\% & 14.0\% & 3.1\% & 0.4\%\tabularnewline
 & Independent Single RNN & 32.3\% & 23.1\% & 15.3\% & 8.9\% & 6.8\% & 6.3\% &  & 37.5\% & 25.0\% & 20.0\% & 12.3\% & 8.3\%\tabularnewline
 & Joint Single RNN & 40.3\% & 25.3\% & 20.8\% & 13.4\% & 8.4\% & 7.6\% &  & 43.7\% & 25.6\% & 21.0\% & 13.5\% & 8.0\%\tabularnewline
 & Synced Pair RNN & 41.4\% & 25.8\% & 19.9\% & 13.4\% & 8.5\% & 7.8\% &  & 41.6\% & 28.6\% & 20.5\% & 13.0\% & 7.9\%\tabularnewline
 & Farha \etal \cite{farha2018when} & \textbf{55.7\%} & \textbf{41.7\%} & \textbf{35.3\%} & \textbf{29.7\%} & \textbf{26.8\%} & \textbf{28.2\%} &  & 46.8\% & \textbf{33.8\%} & \textbf{27.0\%} & \textbf{22.1\%} & \textbf{22.9\%}\tabularnewline
\midrule 
\multirow{5}{*}{Fine} & Dummy & 19.2\% & 4.8\% & 1.3\% & 0.5\% & 0.0\% & 0.0\% &  & 18.8\% & 5.0\% & 1.9\% & 0.2\% & 0.0\%\tabularnewline
 & Independent Single RNN & 21.1\% & 11.8\% & 8.8\% & 5.6\% & 3.8\% & 3.3\% &  & 18.4\% & 8.9\% & 5.7\% & 3.9\% & 2.3\%\tabularnewline
 & Joint Single RNN & 15.8\% & 7.4\% & 5.8\% & 3.6\% & 2.4\% & 2.1\% &  & 14.6\% & 8.3\% & 6.5\% & 3.8\% & 2.2\%\tabularnewline
 & Synced Pair RNN & 22.1\% & 10.4\% & 7.0\% & 4.6\% & 2.6\% & 2.3\% &  & 20.5\% & 8.8\% & 5.4\% & 3.2\% & 1.7\%\tabularnewline
 & Farha \etal \cite{farha2018when} & \textbf{24.8\%} & \textbf{17.7\%} & \textbf{14.4\%} & \textbf{9.8\%} & \textbf{7.5\%} & 7.8\% &  & \textbf{29.7\%} & \textbf{19.1\%} & \textbf{13.8\%} & \textbf{8.7\%} & 7.5\%\tabularnewline
\midrule 
Mid & \multirow{2}{*}{HERA} & 46.8\% & 34.1\% & 24.8\% & 18.9\% & 15.7\% & 19.9\% &  & 41.5\% & 31.3\% & 23.7\% & 16.3\% & 18.9\%\tabularnewline
Fine &  & 21.9\% & 13.1\% & 9.0\% & 5.9\% & 5.3\% & \textbf{8.3\%} &  & 20.5\% & 12.5\% & 9.7\% & 6.4\% & \textbf{8.7\%}\tabularnewline
\bottomrule
\end{tabular}}
\end{table*}

\end{document}